\documentclass[10pt,twocolumn,letterpaper]{article}
\usepackage{multirow}
\usepackage{float}
\usepackage[pagenumbers]{cvpr} % To force page numbers, e.g. for an arXiv version

\definecolor{cvprblue}{rgb}{0.21,0.49,0.74}
\usepackage[pagebackref,breaklinks,colorlinks,allcolors=cvprblue]{hyperref}
 
\usepackage{marvosym}

\newcommand{\vaeshortname}{VidFaceVAE}
\newcommand{\shortname}{VividFace}

\def\paperID{} % *** Enter the Paper ID here
\def\confName{}
\def\confYear{}

\title{VividFace: A Diffusion-Based Hybrid Framework for High-Fidelity \\ Video Face Swapping}

\author{%
Hao Shao$^{1,2}$~~~ Shulun Wang$^{2}$~~~ Yang Zhou$^{2}$ ~~~ Guanglu Song$^{2}$ \\ 
  Dailan He$^{1}$ ~~~ Shuo Qin$^{2}$~~~ Zhuofan Zong$^{1}$~~~ Bingqi Ma$^{2}$~~~ Yu Liu$^{2}$~\textsuperscript{\Letter} ~~~ Hongsheng Li$^{1,3}$~\textsuperscript{\Letter} \\ \\
$^1$CUHK MMLab~~~ $^2$SenseTime Research ~~~ $^3$CPII under InnoHK
}

\usepackage{indentfirst}

\begin{document}

\twocolumn[{
\maketitle

    \vspace{-1cm}
\begin{center}

    \includegraphics[width=1.0\textwidth]{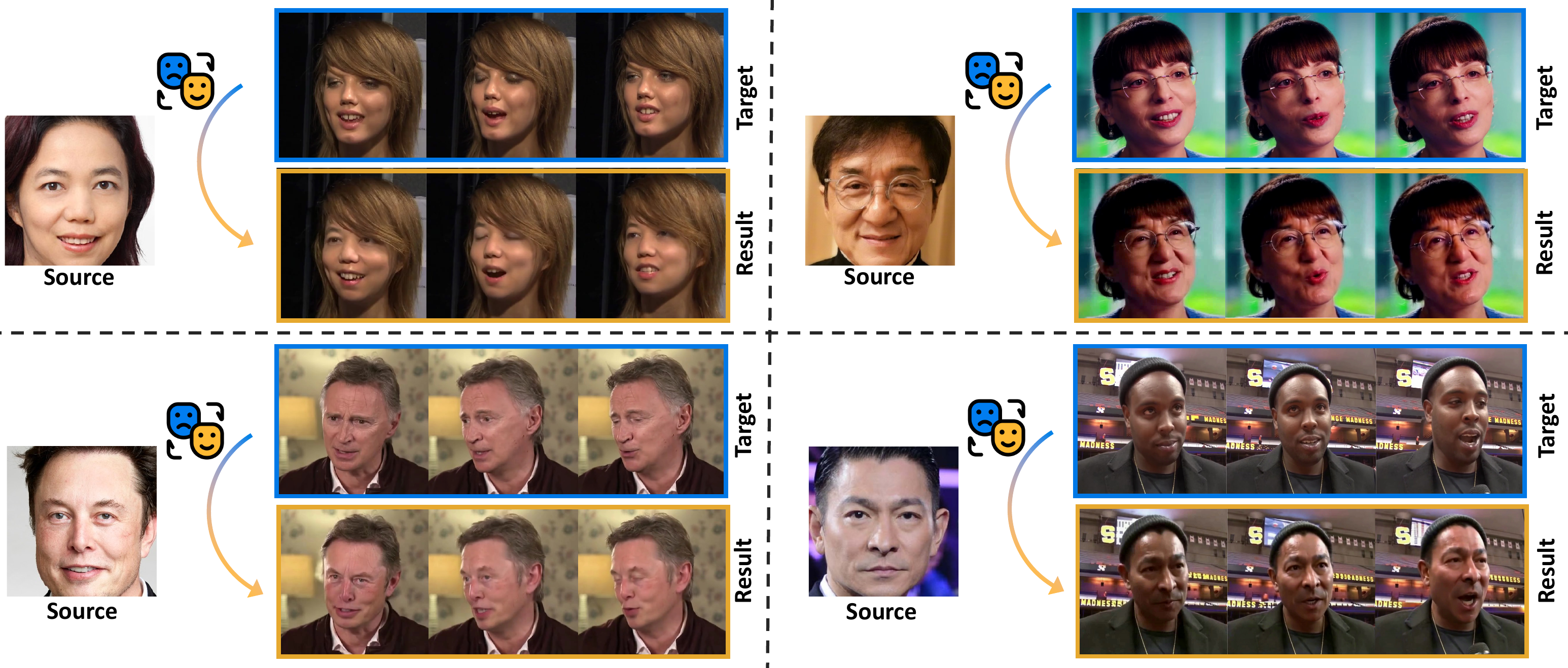}
    \vspace{-0.25cm}
    \captionof{figure}{Face swapping results of \shortname~ at $512 \times 512$ resolution. Our method produces high-fidelity and vivid outputs that accurately follow both pose and expression changes.}
    \label{fig:intro}
\end{center}
}]
    \vspace{-1cm}

\begin{abstract}

\noindent Video face swapping is becoming increasingly popular across various applications, yet existing methods primarily focus on static images and struggle with video face swapping because of temporal consistency and complex scenarios. In this paper, we present the first diffusion-based framework specifically designed for video face swapping. Our approach introduces a novel image-video hybrid training framework that leverages both abundant static image data and temporal video sequences, addressing the inherent limitations of video-only training. The framework incorporates a specially designed diffusion model coupled with a  \vaeshortname\ that effectively processes both types of data to better maintain temporal coherence of the generated videos. 
To further disentangle identity and pose features, we construct the Attribute-Identity Disentanglement Triplet (AIDT) Dataset, where each triplet has three face images, with two images sharing the same pose and two sharing the same identity.
Enhanced with a comprehensive occlusion augmentation, this dataset also improves robustness against occlusions.
Additionally, we integrate 3D reconstruction techniques as input conditioning to our network for handling large pose variations. Extensive experiments demonstrate that our framework achieves superior performance in identity preservation, temporal consistency, and visual quality compared to existing methods, while requiring fewer inference steps. Our approach effectively mitigates key challenges in video face swapping, including temporal flickering, identity preservation, and robustness to occlusions and pose variations. For more information, please refer to the \href{https://hao-shao.com/projects/vividface.html}{project page}.
\end{abstract}    
\section{Introduction}
\label{sec:intro}

\noindent In recent years, face swapping has emerged as a crucial technology across various domains, from content creation~\cite{perov2020deepfacelab} and privacy protection~\cite{wu2019privacy} to safe stunt scene production~\cite{ren2021pirenderer} and digital twin generation~\cite{one2016rogue}. As video is a predominant medium for communication, the demand for high-quality face swapping techniques has grown substantially. Video face swapping involves extracting identity features from a source face and seamlessly integrating them with the attributes (such as expressions, poses, \textit{etc.}) and background of a target face while maintaining temporal consistency. However, despite the recent advancements, current face-swapping methods encounter difficulties in video contexts, as most are optimized for static images rather than dynamic video sequences.

Existing face swapping approaches can be broadly categorized into three main methodologies: 3D-based, GAN-based, and diffusion-based methods. Traditional 3D-based methods~\cite{bitouk2008face,blanz2004exchanging,thies2016face2face,nirkin2018face}, primarily utilizing 3D Morphable Models (3DMM)~\cite{blanz2023morphable}, often struggle with low-resolution outputs and face blending issues. GAN-based approaches~\cite{nirkin2019fsgan,chen2020simswap,li2019faceshifter,bao2018towards,liu2023fine,zhu2021one}  encounter challenges with training instability, mode collapse, and producing low-resolution output, particularly in complex cases. Recently, diffusion models~\cite{ho2020denoising} have gained prominence in image synthesis tasks, offering advantages such as high-fidelity output, enhanced controllability, and high training stability.

Although recent works like DiffSwap~\cite{zhao2023diffswap} and REFace~\cite{baliah2024realistic} have demonstrated the potential of diffusion models in image-level face swapping, significant challenges remain unaddressed, particularly for face swapping in videos. 
The video domain introduces additional complexities, including temporal consistency maintenance, handling large pose variations, and addressing occlusions.
To tackle these challenges, we propose the first diffusion-based video face swapping framework \shortname~ that adopts a novel image-video hybrid training strategy. This approach addresses the inherent limitations of video-only training, such as limited diversity of videos where hundreds of frames in a single video tend to be highly similar. By incorporating abundant and readily accessible image data into the training process, our framework significantly enhances the diversity of training samples. Our framework incorporates a specifically designed diffusion model that can process both static images and temporal video data, coupled with a \vaeshortname. Unlike the standard VAE~\cite{kingma2013auto}, our \vaeshortname ~is uniquely designed to process both face images and videos in a unified embedding space and was trained with a large-scale face dataset. It makes image data bypass motion-related layers while adaptively merging outputs from both image and video branches, effectively mitigating temporal flickering and text disorders commonly encountered in video face swapping.

To enhance the identity similarity of swapped faces, we introduce the Attribute-Identity Disentanglement Triplet (AIDT) Dataset. Each data triplet consists of a source face, a target face, and a GAN-generated decoupling face. The source and target faces share the same identity but differ in pose and expression, while the GAN-generated face matches the target face’s pose and expression yet features a different identity. This dataset design improves the model's ability to disentangle identity and pose features, enhancing both identity and attribute preservation. Additionally, the proposed occlusion data augmentation strategy adds various types of occluding objects to partially cover the faces with dynamic temporal patterns to the target images. To further address large pose variations, we incorporate 3D reconstruction of the target face as a conditioning input, using a 3D Morphable Model (3DMM). This 3D guidance helps the diffusion model accurately capture the pose and expression of the swapped face, promoting generalization across diverse videos.

Experimental results demonstrate our framework's superiority in terms of Fréchet Video Distance (FVD), temporal consistency, and attribute/identity preservation, with fewer inference steps compared to existing methods. Besides, we also demonstrate the stability and generation of our method in multiple complex cases.

To summarize, this paper makes the following contributions:
\begin{itemize}
\item We propose the first diffusion-based framework \shortname~, specifically designed for video face swapping, featuring a novel image-video hybrid training strategy that effectively leverages both static image and temporal video data.
\item We introduce a PIDT dataset construction method that enhances identity-expression disentanglement, along with a comprehensive occlusion augmentation strategy that improves robustness to real-world scenarios.
\item We effectively integrate 3D face reconstruction techniques and use the rendered output as additional conditioning information, enabling our framework to handle challenging conditions such as large pose variations while avoiding potential shortcuts in the generation process.
\item Extensive experiments demonstrate our framework's superior performance in terms of temporal consistency, identity preservation, and visual quality. Our ablation studies and component analysis provide valuable insights for future research in video face swapping.
\end{itemize}

\section{Related Work}
\label{sec:related_work}
\subsection{Face Swapping}

\noindent The frameworks of face swapping are generally categorized into three types: 3D-based~\cite{bitouk2008face,blanz2004exchanging,thies2016face2face,nirkin2018face}, GAN-based~\cite{nirkin2019fsgan,chen2020simswap,li2019faceshifter,bao2018towards,liu2023fine,zhu2021one}, and diffusion-based methods~\cite{kim2022diffface,zhao2023diffswap,baliah2024realistic}. 3D-based frameworks typically employ the parameterized 3DMM~\cite{blanz2023morphable} model to reconstruct the swapped face. 
Face2Face~\cite{thies2016face2face} transferred expressions from source to target face by fitting a 3DMM face model to both faces.
The authors in \cite{nirkin2018face} show that face swapping with robust segmentation preserves identity in intra-subject swaps and reduces recognizability in inter-subject cases.
HifiFace~\cite{wang2021hififace} additionally designed a semantic facial fusion module for adaptive feature blending to make the results more photo-realistic.
However, these 3D-based methods exhibit low similarity with the source images and suffer from unrealistic texture and lighting due to the limited resolution of 3D face models.

GAN~\cite{goodfellow2014generative} has been a powerful tool for generating realistic synthetic images.
The popular algorithm DeepFakes~\cite{DeepFakes2020faceswap} utilizes an encoder-decoder architecture for identity-specific face swapping but lacks generalization. To improve adaptability, FSGAN~\cite{nirkin2019fsgan} proposes a subject-agnostic approach with a recurrent reenactment module, inpainting and a blending module.
SimSwap~\cite{chen2020simswap} extends the flexibility with an ID Injection Module and Weak Feature Matching Loss, achieving high-fidelity results for arbitrary identities. 
FaceShifter~\cite{li2019faceshifter} introduces a two-stage framework with an Adaptive Embedding Integration Network and a refinement module.
However, GAN-based methods often struggle with balancing multiple loss functions and managing large shape differences or occlusions, leading to challenges in maintaining illumination consistency and identity fidelity in complex scenarios.

Recently, diffusion models have become a leading framework for image \& video generation. 
Face swapping can also be formulated as an inpainting task, leveraging the identity and attribute features as condition vectors. 
DiffFace~\cite{kim2022diffface} first introduces a diffusion-based framework with ID Conditional DDPM and target-preserving blending for stable, high-fidelity face swapping.
DiffSwap~\cite{zhao2023diffswap} employs a powerful diffusion model that follows the conditional inpainting paradigm to generate high-fidelity and controllable swapped faces.
REFace~\cite{baliah2024realistic} further advances this approach by reframing face-swapping as a self-supervised inpainting task, improving identity transfer and fidelity with minimal inference time. However, these methods are limited to static images and do not address video face-swapping, which requires temporal consistency and more robust handling of large poses and occlusions.

\subsection{Diffusion Models}

\noindent Diffusion models~\cite{ho2020denoising} have recently emerged as a powerful generative framework, achieving state-of-the-art performance in various domains, including image synthesis~\cite{dhariwal2021diffusion, ho2020denoising, song2020denoising}, editing~\cite{kawar2023imagictextbasedrealimage, shi2024dragdiffusion, lin2024schedule}, super-resolution~\cite{wang2024sinsr,gao2023implicit}, and video generation~\cite{guo2023animatediff, blattmann2023stable, ma2024latte}. Unlike GANs, which often suffer training instability, diffusion models offer a more stable training process by gradually denoising data from random noise, resulting in high-fidelity outputs. Notable advancements include Stable Diffusion~\cite{rombach2022high}, which enhances efficiency by operating in the latent space, and SVD~\cite{blattmann2023stable}, which incorporates temporal modules to scale diffusion models for video tasks. Conditioning mechanisms, such as cross-attention and concatenation, further improve controllability, enabling targeted generation across applications~\cite{tian2024emo}. With these advantages, diffusion models have become increasingly popular for high-quality, versatile content creation.

\section{Method}
\label{sec:method}

\begin{figure*}[]
    \centering
    \includegraphics[width=0.95\textwidth]{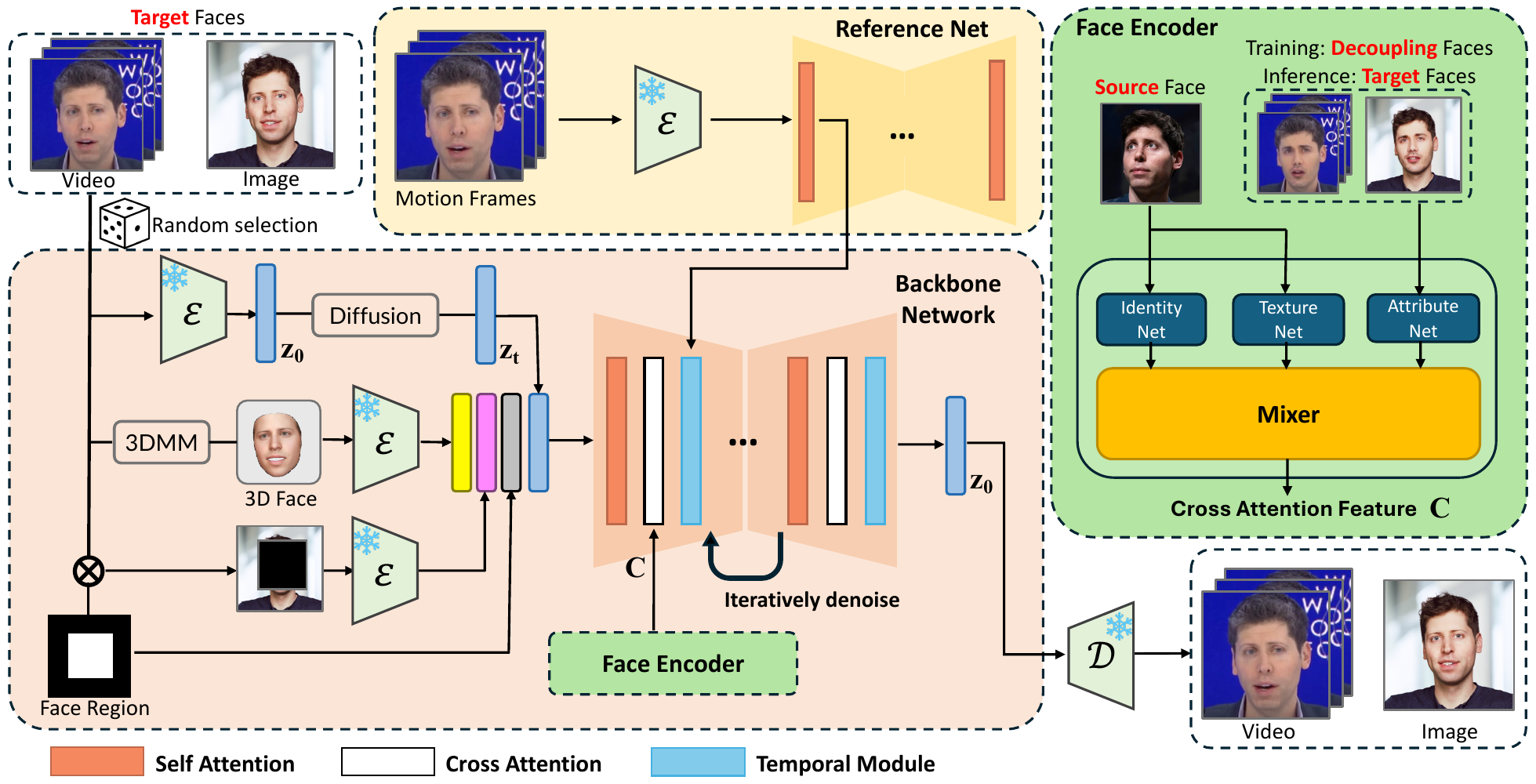}
    \caption{Overview of the proposed framework. During training, our framework randomly chooses static images or video sequences as the training data. In addition to the noise $z_t$, three other types of inputs are integrated to guide the generation process: (1) a face region mask, which controls the generation of facial imagery; (2) a 3D reconstructed face, which helps guide the pose and expression, especially in cases of large pose variations; and (3) masked source images, which supply background information. These inputs are processed through the Backbone Network, which performs the denoising operation. Within the Backbone Network, we employ cross-attention and temporal attention mechanisms. The temporal attention module ensures temporal continuity and consistency across frames. Our face encoder extracts identity and texture features from the target face, as well as pose and expression details from the source face, and uses these features in cross-attention to produce realistic and high-fidelity results. }
    \label{fig:framework}
\end{figure*}

\noindent In this section, we will introduce our method \shortname, the first diffusion model based video face swapping framework in detail. An overview of our framework is shown in Fig.~\ref{fig:framework}.

\subsection{Preliminaries}
\noindent Our method employs Stable Diffusion (SD)~\cite{rombach2022high} as the backbone network. Stable Diffusion is a text-to-image model built on the Latent Diffusion Model (LDM), which enables efficient image generation by operating within a compressed latent space. SD uses a variational autoencoder (VAE)~\cite{kingma2013auto} to map the original image $x_{0}$ unto a latent representation $z_{0}$. reducing computational cost while preserving visual quality. The image is encoded as $z_{0} = \mathcal{E}(x_{0})$ and decoded back as $x_{0} = \mathcal{D}(z_{0})$. SD follows the Denoising Diffusion Probabilistic Model (DDPM)~\cite{ho2020denoising} framework, introducing Gaussian noise $\epsilon$ to the latent $z_{0}$ across timesteps $t$, generating a noisy latent $z_{t}$ over a series of steps. During inference, the model denoises $z_{t}$ back to $z_{0}$, guided by condition features. The denoising backbone $\epsilon_{0}$, based on a U-Net~\cite{ronneberger2015u}, is trained to predict the noise and remove it progressively, using the objective:

$$
L = \mathbb{E}_{t, c, z_t, \epsilon} \left[ \| \epsilon - \epsilon_\theta(z_t, t, c) \|^2 \right],
$$
where $c$ represents text features derived from a CLIP encoder~\cite{radford2021learning}. SD uses a U-Net with cross-attention mechanisms, to fuse text embeddings with latent features, enabling fine control over generated images based on text prompts. This allows SD to generate detailed, high-fidelity images while responding effectively to user input.

% Animatidiff

\subsection{Hybrid Face Swapping Framework}

\noindent \textbf{Video Face Swapping Task.} The goal of video face swapping is to transfer the identity of a source face onto a target video while preserving the target’s pose, expression, lighting, and background. Following the approach of DiffSwap~\cite{zhao2023diffswap}, we model the video face swapping task as conditional inpainting. This involves masking the face region in the target frame and injecting conditioning vectors that represent the identity of the source face and the attributes of the target face to guide the generation process. Recent diffusion-based methods~\cite{zhao2023diffswap, kim2022diffface, baliah2024realistic} have primarily focused on static image face swapping. However, directly applying these methods to video sequences by decomposing the video into multiple static images introduces new problems, such as temporal distortions, flickering, occlusions, and issues with large pose variations.

To address these challenges, we propose an image-video hybrid framework, \shortname~for video face swapping. Our framework also incorporates image-level training data to enhance face swapping performance. Specifically, we first train a \vaeshortname~ that can transform source images $x_{src}^{i} \in \mathbb{R}^{1 \times 3 \times H \times W}$ or videos $x_{src}^{v} \in \mathbb{R}^{T \times 3 \times H \times W}$ into the latent space $z_{0}  \in \mathbb{R}^{T \times C \times H \times W}$, where $H$ and $W$ indicate the spatial dimensions of the input data, $T$ is the count of generated frames, and $C$ is the feature dimension.
In our framework, static images are treated as single-frame videos. We then train a conditional diffusion model $\epsilon_{\theta}(z_{t},t; \textbf{C})$ to perform denoising on that latent space with a specific emphasis on identity consistency, where $\textbf{C}$ denotes the conditioning vectors and $t$ denotes the denoising timestep.
During training, as ground truth is unavailable when the source and target images come from different individuals, we utilize pairs of face images from the same individual as source-target training pairs. 
As illustrated in Figure~\ref{fig:framework}, our framework randomly selects either static images or video sequences as training data. To synchronize gradients, each batch contains only one type of data.

\begin{figure}[]
    \centering
    \includegraphics[width=0.99\linewidth]{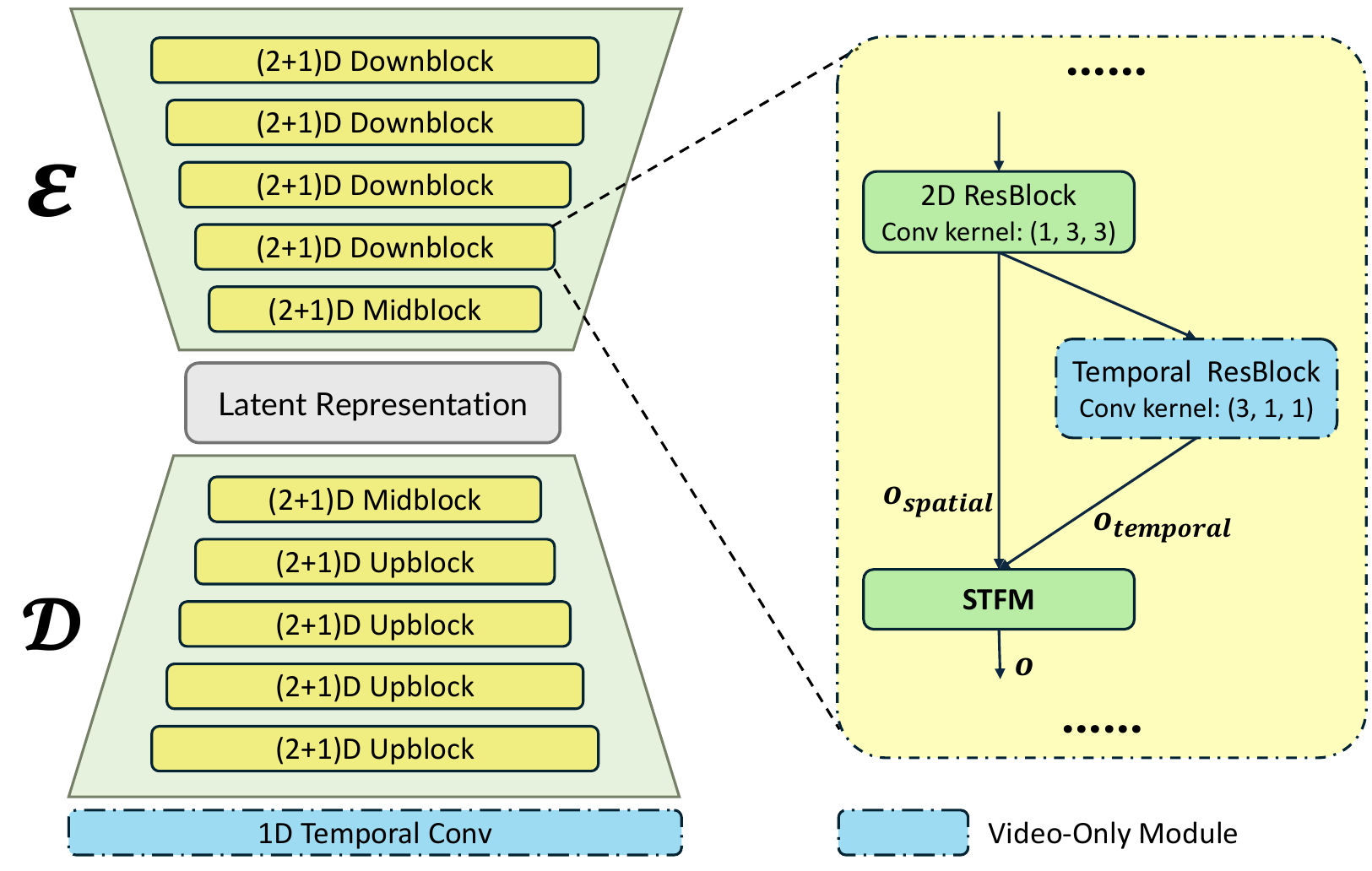}
    \vspace{-0.3cm}
    \caption{Overview of the proposed \vaeshortname, capable of simultaneous encoding and decoding of both image and video data. Certain modules are specifically designed for video inputs, and image inputs bypass these modules as needed.}
    \label{fig:3dvae}
        \vspace{-0.3cm}
\end{figure}

\noindent \textbf{\vaeshortname.} As shown in Figure~\ref{fig:3dvae}, our proposed VidFaceVAE is a VAE framework designed to enhance the reconstruction quality of facial data, effectively handling both video sequences and static images. The \vaeshortname~ primarily consists of (2+1)D blocks, combining 2D spatial and 1D temporal convolutions to form pseudo-3D operators. For image inputs, the STFM (Spatial Temporal Fusion Module) outputs the result of the 2D ResBlock directly, bypassing the temporal ResBlock. For video inputs, the STFM combines the outputs from both the 2D and temporal blocks using a learnable coefficient $\beta$, described as $o = \beta \times o_{\text{spatial}} + (1- \beta) \times o_{\text{temporal}},$
where $o_{spatial}$ and $o_{\text{temporal}}$ denote the output from the spatial branch and the temporal branch. We do not involve the temporal downsampling modules in our VAE framework as it needs to process image data.
The \vaeshortname~ employs a (2+1)D structure with two primary advantages: (1) It reduces the computational cost by decoupling spatial and temporal convolutions, making it more efficient than full 3D convolutions. (2) It enables the reuse of pretrained 2D VAE parameters and SD pretrained weights to accelerate convergence and improves the final performance.
Unlike OD-VAE~\cite{chen2024od}, we avoid the 3D-Causal-CNN, as temporal modules are skipped from images and our backbone network is not based on transformers. Causal convolution limits the model's capacity and using casual conv processing static images can not bring performance improvement for both videos and images.

\noindent \textbf{Temporal Modules.}  Inspired by the architectural concepts of AnimateDiff~\cite{guo2023animatediff}, EMO~\cite{tian2024emo}, and V-Express~\cite{wang2024v}, we incorporate self-attention temporal layers into the features within frames. In our framework, temporal modules are exclusively applied to video sequences. For training with video inputs, we prepare motion frames $x_{src}^{motion} \in \mathbb{R}^{M \times 3 \times H \times W}$, sampled from clips preceding the source videos, which are fed into the ReferenceNet to extract feature maps following each self-attention module.
During the denoising process in the Backbone Network, we merge the temporal layer inputs with the pre-extracted motion features at matching resolutions along the frame dimension. The temporal dimension of these concatenated feature maps becomes $M+T$, enabling the application of temporal attention. This design enhances temporal coherence across frames. During training, we randomly initialize the motion frames as zero vectors to animate the generation of the first video clip. 

\begin{figure}[]
    \centering
    \includegraphics[width=0.45\textwidth]{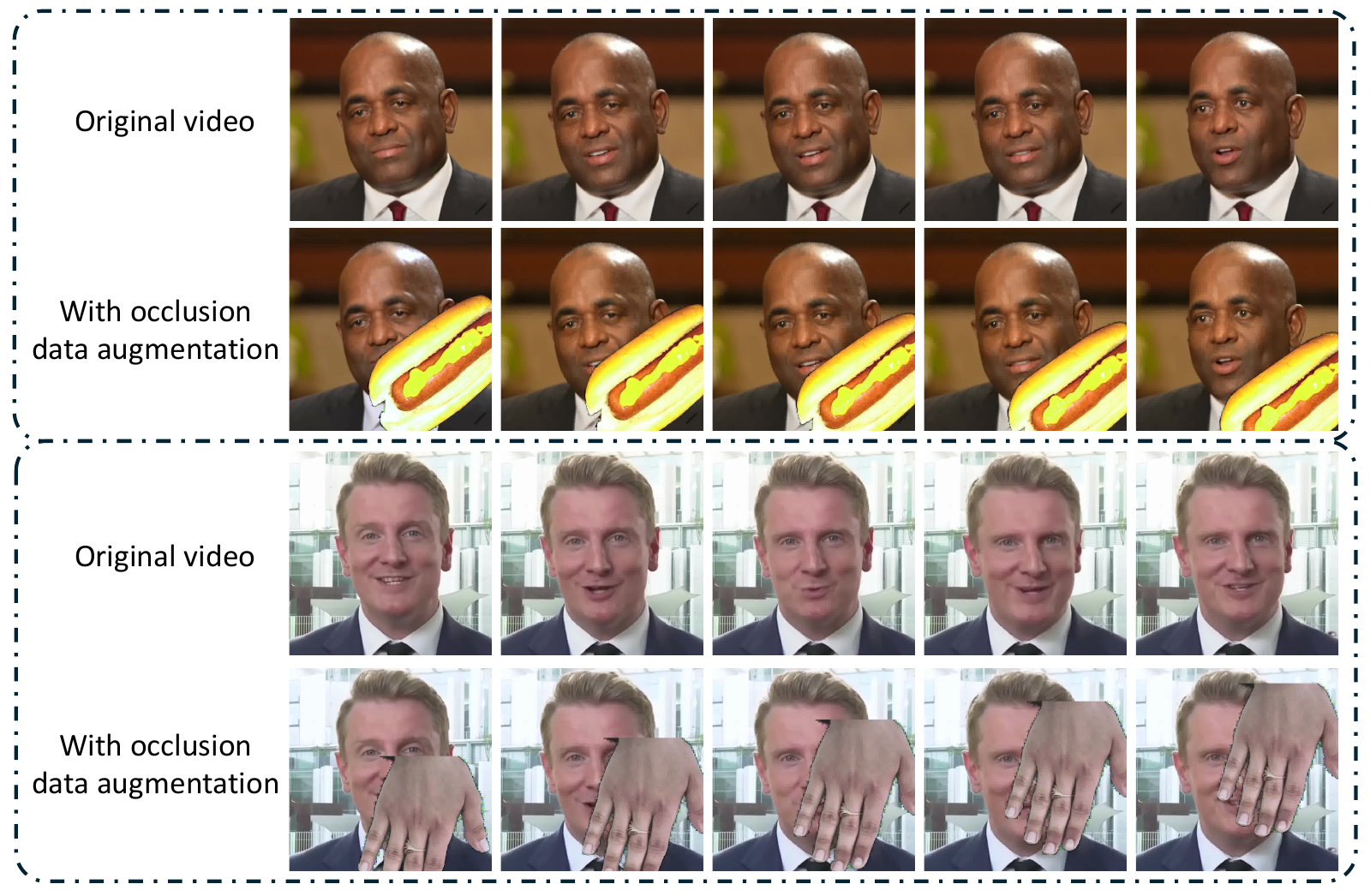}
    \vspace{-0.3cm}
    \caption{Visualization of our occlusion data augmentation, which improves the stability and consistency of the generated videos.}
    \vspace{-0.3cm}
    \label{fig:occlusion}
\end{figure}

\vspace{-0.15em}
\subsection{Designs of Condition Vectors}
\vspace{-0.15em}

\noindent In our framework, several carefully designed condition vectors are used to guide the generation process, ensuring accurate and consistent visual outputs for both static images and video sequences. We formulate video face swapping as a conditional inpainting task, where masked videos with cropped face regions provide the background and lighting conditions. The corresponding face regions guide the diffusion model on where are generated the faces. 

In many in-the-wild videos, faces often exhibit significant pose variations, which can lead diffusion models to produce suboptimal results, such as facial distortions and inaccurate pose estimations.
To address this issue, we propose using a 3D reconstruction technique to reconstruct the face and use its output as local guidance for pose and expression details. Specifically, we employ 3DMM~\cite{blanz2023morphable} to extract BFM coefficients, setting the texture component to zero to reduce information leakage.
Replacing the reconstructed face with the original face would introduce even more pronounced information leakage. Since the ground truth face is identical to the input target face, excessive information leakage would degrade the model's generalization capabilities.
To ensure that the generated face maintains the same identity as the source face while preserving attributes (such as pose, expression, \textit{etc.},), we inject cross-attention features $\textbf{C}$ extracted by our face encoder as global context to the diffusion model.

\begin{figure}[]
    \centering
    \includegraphics[width=0.45\textwidth]{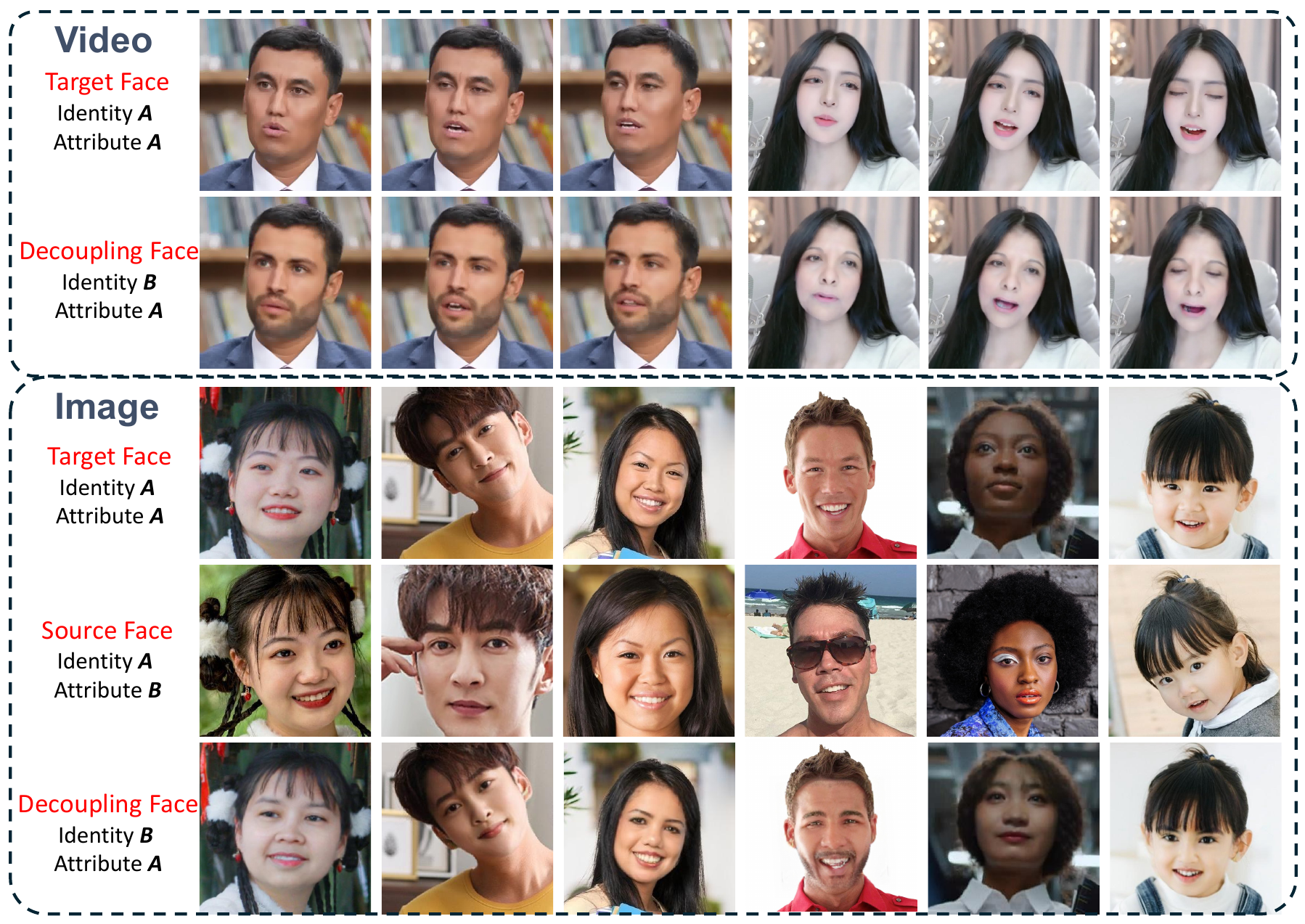}
        \vspace{-0.3cm}
    \caption{Visualization of our AIDT dataset. For video facial data, we present only the target and decoupling faces, as the source faces can be derived from any other frame within the same video clip.}
        \vspace{-0.3cm}
    \label{fig:aidt}
\end{figure}

\noindent \textbf{Face Encoder.} The face encoder module in our framework plays a critical role in extracting and integrating features from the target and source faces to guide the face-swapping process effectively. As illustrated in the right part of Figure~\ref{fig:framework}, the face encoder is composed of three primary networks, each responsible for capturing distinct aspects of facial information: (1) identity net: this network focuses on extracting the core identity features from the target face; (2) texture net: this network is designed to capture detailed texture information from the target face, such as skin tone, fine facial features; (3) attribute net: the net extracts additional facial attributes from the source, such as pose, expression, and other dynamic features that contribute to a realistic and expressive representation.

The straightforward approach is to send the source image to both the identity and texture networks, while the target image is sent to the attribute network. However, a challenge arises when the source and target faces do not belong to the same person, as the ground truth is unavailable in the real world. In most previous methods~\cite{baliah2024realistic, zhao2023diffswap, kim2022diffface}, the source and target images are assumed to be the same, meaning all three networks receive identical input. This results in difficulties for the face encoder in extracting distinct features and leading to information leakage. Specifically, this leakage causes the model to merely ``copy and paste" the face region, effectively completing the task by superficially transferring facial features without meaningful feature disentangling or transformation.
In contrast, our framework, built on the AIDT dataset (shown in Figure~\ref{fig:aidt}), employs source images (same identity, but different attributes) and decoupling images (same attribute, but different identity). These images help the face encoder disentangle and fuse different components of facial features, thus improving generalization when the source and target faces come from different individuals during inference.
Details of the dataset construction can be found in Sec.~\ref{sec:triplet}.

These extracted features are then passed through an attention mechanism within the Mixer module, where they are multiplied by corresponding weight coefficients before being fused. This process combines the identity, texture, and attribute features to create a comprehensive cross-attention feature representation \textbf{C}. This fused representation provides rich contextual information to guide the diffusion model during face generation, ensuring both high fidelity and identity consistency in the swapped face across video frames.

\subsection{Training Strategies}

\noindent Our training process involves three stages to progressively enhance model performance for video face swapping. 
The first stage focuses on training the \vaeshortname~, where we apply reconstruction, perceptual, and KL divergence losses to ensure high-quality reconstruction and a well-structured latent space.  The training data primarily consists of facial images and videos. Given the specifically designed architecture, the spatial modules are initialized using the original 2D VAE. In subsequent stages, the VAE is frozen and no longer updated.
In the second stage, we pretrain the model using image data, while the ReferenceNet and temporal modules of the backbone network remain inactive. The backbone is initialized from the original SD weights.
The final stage introduces image-video hybrid training, which incorporates temporal coherence by activating the temporal modules and utilizing video data. The temporal modules are initialized from AnimateDiff~\cite{guo2023animatediff}, enabling smooth frame transitions and reducing flicker artifacts.

\section{AIDT Dataset}
\label{sec:triplet}

\noindent In this section, we describe the construction of triplet pairs for our AIDT (Attribute-Identity Disentanglement Triplet) dataset, as illustrated in Figure~\ref{fig:aidt}. 
For image data, we first cluster the facial images based on identity similarity. From each cluster, we randomly select two images to form a target-source pair that shares the same identity but has different attributes.
To generate the decoupling image, which has a different identity but the same attribute, we use the InsightFace Swapper to create synthetic images with a distinct identity, while preserving the gender of the original face.
We have observed that when the original and swapped faces belong to different genders, the results tend to degrade. Additionally, we exclude triplets with significant facial expression discrepancies by comparing the face landmarks.
For video data, the process is similar, except that both the source and target images come from the same video clip, but not from the same frames as the target or motion images, which reduces the pose variation. Since video data is less abundant than image data, clustering does not yield enough pairs to form a sufficient number of triplets.

The AIDT dataset enables the face encoder to disentangle and fuse distinct facial components—ID features, texture features from the source face, and attribute features from the decoupling face. This enhances generalization, especially when the source and target faces belong to different individuals during inference.

\begin{figure*}[]
    \centering
    \includegraphics[width=1\linewidth]{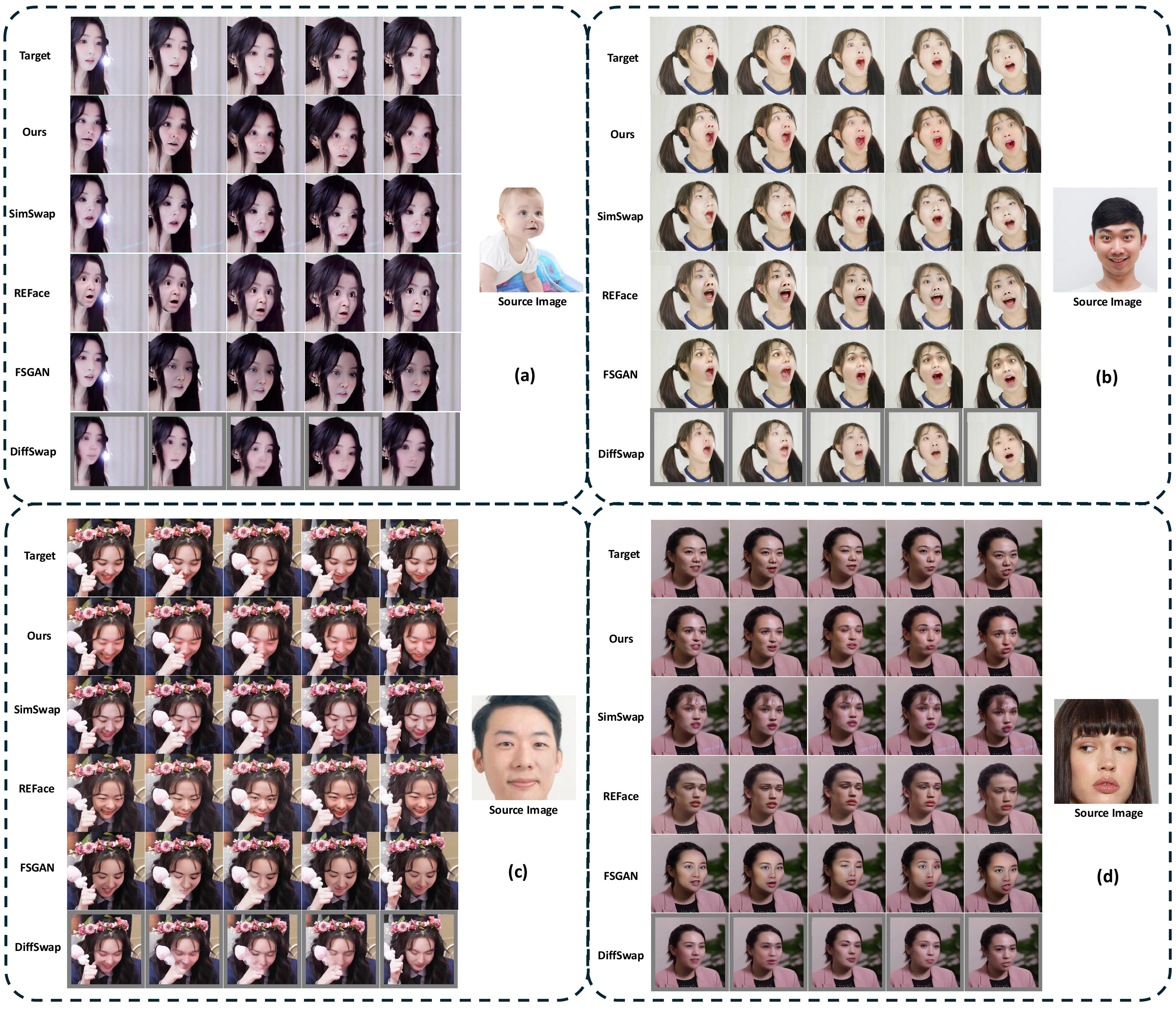}
    \vspace{-0.cm}
    \caption{Qualitative comparison at $512 \times 512$ resolution. Our method generates high-fidelity results and handles challenging cases effectively, such as large poses (b) and occlusions (c). Corresponding videos are provided in the supplementary material.It is best viewed at a larger scale for optimal evaluation.}
    \vspace{-0.cm}
    \label{fig:compare}
\end{figure*}

\section{Experiment}
\label{sec:experiment}

\subsection{Implementation Details}

\noindent We collected approximately 300 hours of facial videos from the internet to train our models, and the facial images are partially sourced from VGGFace2-HQ~\cite{chen2023simswap++}.
In our experiments, we use a latent space of size $13 \times 64 \times 64$ and a U-Net architecture for the $\epsilon_0$ denoising network. Images and video clips sampled from the dataset are resized and cropped to $ 512 \times 512 $. The number of motion frames, $M$, is set to 4, and the generated video length, $T$, is set to 8 frames.
For the face encoder, the identity network is based on ArcFace~\cite{deng2019arcface}, while the texture and attribute networks are based on DINO~\cite{caron2021emerging}. We use SCRFD~\cite{guo2021sample} for facial bounding box detection. The AdamW optimizer~\cite{loshchilov2017decoupled} is used for training.
In the first stage of the VAE training, the learning rate is set to 5e-6 with a batch size of 32. The weights of reconstruction, perceptual, and KL divergence loss are 1.0, 0.1, 1e-6 respectively. For the second and third stages, the learning rate is increased to 1e-5, with the batch size remaining at 32. During inference, we generate video clips using the DDIM sampling algorithm for 32 steps.

\subsection{Evaluation Protocol}

\noindent Considering that most previous baselines, such as CelebA~\cite{lee2020maskgan} and FFHQ~\cite{karras2019style}, are primarily focused on image face swapping, we propose a new benchmark for video face swapping. Our benchmark includes 200 source images and 200 high-resolution target videos, with each video containing 128 frames and a single trackable face. These videos and images feature unseen identities and backgrounds, ensuring a diverse and challenging dataset.
To evaluate performance, we generate 200 swapped videos using our framework. For comparison, since other methods are based on image-level face swapping, we perform face swapping frame by frame for those methods.
For facial data reconstruction, we use SSIM, PSNR and LPIPS~\cite{zhang2018unreasonable} to evaluate the quality of reconstructed images and videos. For video face swapping, we use FVD~\cite{unterthiner2019fvd} to assess the overall quality of the generated videos. The attribute transfer error is measured by pose and expression errors. We use HopeNet~\cite{doosti2020hope} and Deep3DFaceRecon~\cite{deng2019accurate} to detect these attributes, and the L2 distance to the ground truth is used as the evaluation metric.
For ID retrieval, we extract identity features from the source images using ArcFace, and for each swapped video, we perform face retrieval by searching for the most similar faces among all source images. The retrieval is measured by the average cosine similarity of all frames, and we report the Top-1 and Top-5 accuracy.

\begin{table}[!htb]
\centering
\setlength{\tabcolsep}{2pt}
\resizebox{0.95\linewidth}{!}{
\begin{tabular}{lcccccc}
\toprule
\multirow{2}{*}{\textbf{Method}}  & \multirow{2}{*}{\textbf{FVD$_{32}$}$\downarrow$}    &   \multirow{2}{*}{\textbf{FVD$_{128}$}$\downarrow$}    & 
 \multicolumn{2}{c}{\textbf{ID retrieval $\uparrow$}}   & \multirow{2}{*}{\textbf{Pose}$\downarrow$}   &  \multirow{2}{*}{\textbf{Expr.}$\downarrow$}    \\ 
 & & &  Top-1 &  Top-5& & \\
\midrule
SimSwap~\cite{chen2020simswap}& \underline{1242.8} & \underline{186.6} & \underline{76.5}  & \underline{88.5}& \textbf{5.12}  &  0.76    \\
FSGAN~\cite{nirkin2019fsgan} &1507.9 & 423.8 &  24.5 & 40.0  &   \underline{5.19} & \underline{0.73}    \\
DiffFace~\cite{kim2022diffface} & 2404.7 & 1404.9 & 1.5  & 4.1  &  18.3 & 1.58   \\
DiffSwap~\cite{zhao2023diffswap} & 1530.2 & 809.3 & 14.5 & 26.3 & 12.9  &  1.02   \\
REFace~\cite{baliah2024realistic} & 1336.9 &  311.9 & 71.9  & 86.5 & 6.67 & 0.91 \\
 \midrule
\textbf{\shortname}~ & \textbf{1201.1} & \textbf{122.6} & \textbf{78.3}  & \textbf{90.2} & 5.43 & \textbf{0.72}     \\
\bottomrule
\end{tabular}}
\vspace{-0.cm}
\caption{Qualitative Comparison. Best is in bold and second best is underlined. our method achieves very competitive results compared with existing methods.} 
\vspace{-0.cm}
\label{tab:comp}
\end{table}

\subsection{Comparisons with Existing Methods}

\noindent \textbf{Qualitative Results.} Since videos cannot be displayed in the PDF and due to submission policy restrictions on showing generated videos, we provide several comparison videos in the supplementary materials and strongly encourage the reader to view them. 
We perform quantitative comparison at $512 \times 512$ resolution. As shown in Figure \ref{fig:compare} (a) and (d), our method generates high-fidelity swapped faces, with attributes that closely match the target faces. In Figure \ref{fig:compare} (b), our method successfully transfers both face shape and expression under large pose variations, benefiting from the 3D reconstruction mask, while other methods exhibit generation artifacts. In Figure \ref{fig:compare} (c), where a toy and hand occlude the girl's face, most other methods fail to handle the occlusion properly, with the toy and hand either displaced or fused together. Additionally, many methods result in noticeable facial deformations. In contrast, our method successfully recovers the occluded areas and maintains accurate face swapping, thanks to our augmentation strategy.

\begin{table}[]
\centering
\resizebox{0.95\linewidth}{!}{
\begin{tabular}{ccccc}
\toprule
\multicolumn{2}{c}{\textbf{Architecture}} & \multicolumn{3}{c}{\textbf{Facial videos}} \\
   \textbf{Encoder} & \textbf{Decoder}         & \textbf{SSIM} $\uparrow$      & \textbf{PSNR} $\uparrow$    & \textbf{LPIPS} $\downarrow$    \\ \hline
2D & 2D                   & 0.967    &        37.61    &     0.048       \\ 
2D & (2+1)D &           0.976   &      38.77      &     0.039       \\ 
(2+1)D & (2+1)D           &  \textbf{0.983}    &      \textbf{41.11}      &  \textbf{0.027}        \\ \bottomrule
\end{tabular}}

\vspace{-0.cm}
\caption{Comparison of different VAE architectures.}
\vspace{-0.cm}
\label{tab:vae}
\end{table}

\noindent \textbf{Quantitative  Results.} In Table~\ref{tab:comp}, we compare five open-source methods (two GAN-based and three diffusion-based). The results show that our method outperforms others in ID retrieval and FVD, generating high-fidelity swapped face videos while preserving the source identity. It also achieves comparable performance in pose and expression, maintaining target attributes effectively.

\begin{figure}[]
    \centering
    \includegraphics[width=0.9\linewidth]{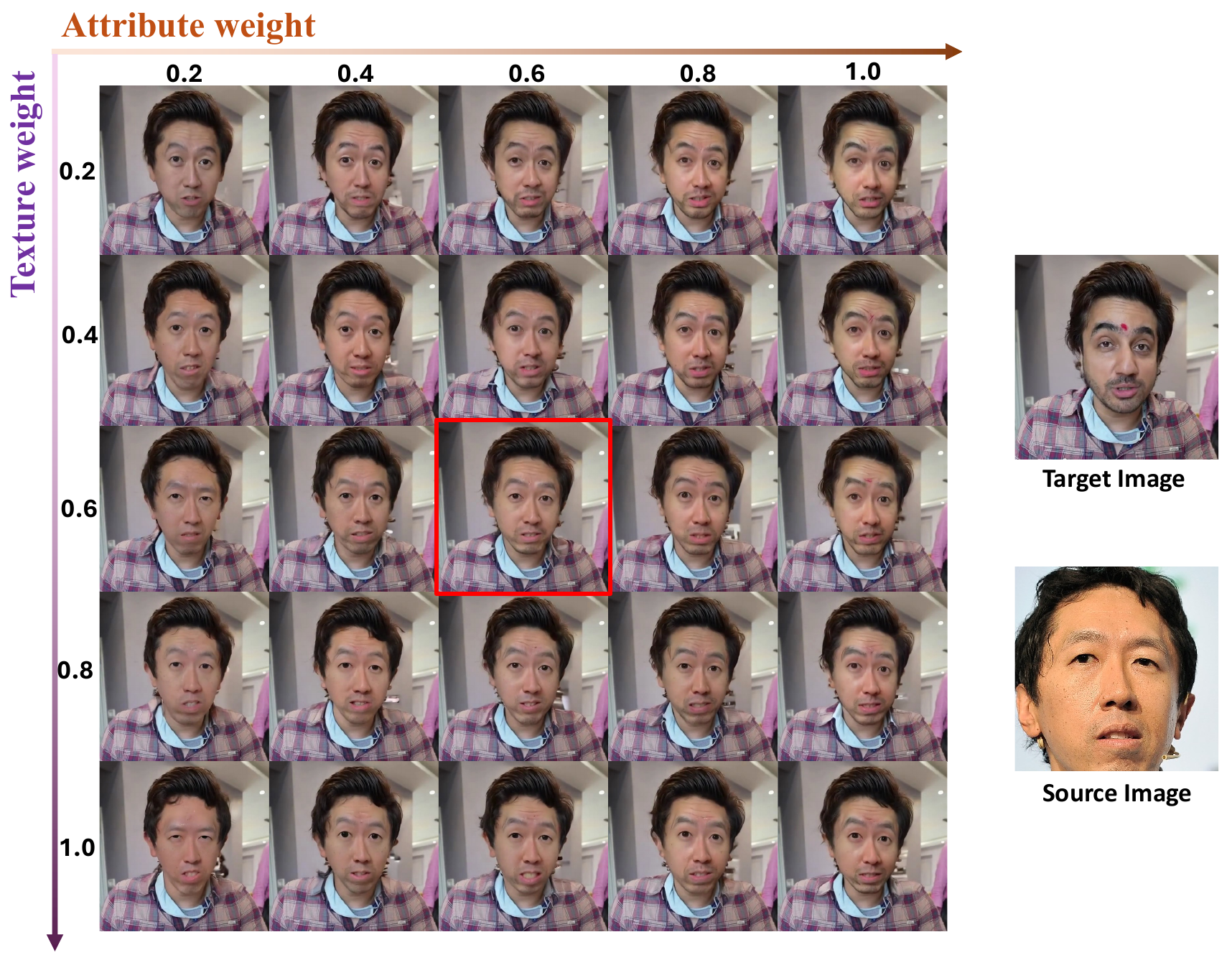}
    \vspace{-0.3cm}
    \caption{Ablation on the different combinations of texture weights and attribute weights.}
    \vspace{-0.3cm}
    \label{fig:mixing}
\end{figure}

\begin{figure}[]
    \centering
    \includegraphics[width=0.9\linewidth]{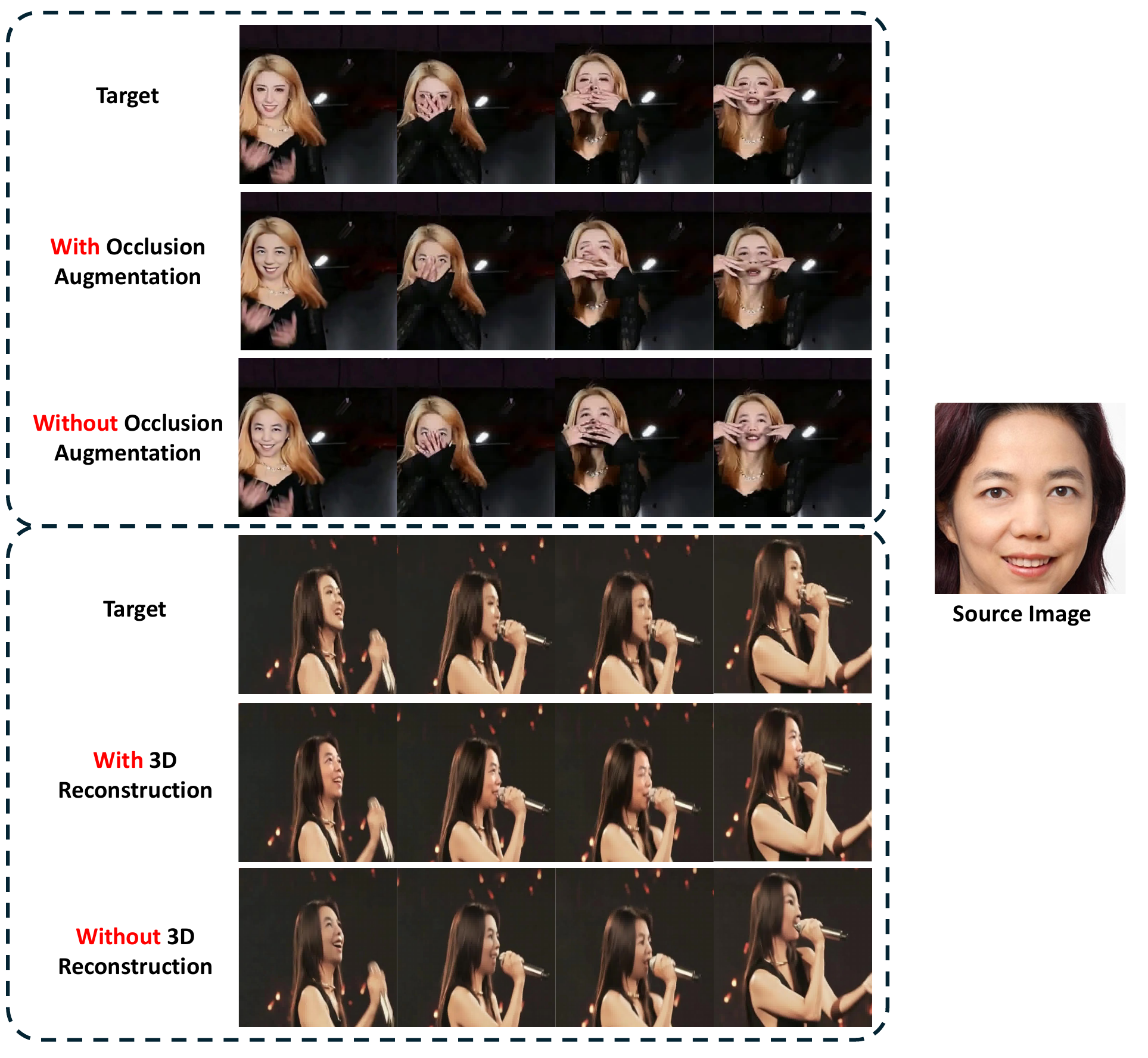}
    \vspace{-0.3cm}
    \caption{Ablation on the occlusion data augmentation and 3D face reconstruction.}
    \vspace{-0.3cm}
    \label{fig:ablation}
\end{figure}

\subsection{Analysis}

\noindent\textbf{VAE architecture.} Table~\ref{tab:vae} shows the reconstruction performance of different VAE architectures. The first model is a pure 2D VAE (SD-VAE), the second uses a (2+1)D decoder with a 2D encoder, and the third is our proposed \vaeshortname~, which uses a full (2+1)D encoder-decoder.
The \vaeshortname~ outperforms the others in all metrics, achieving the highest SSIM (0.983), PSNR (41.11), and the lowest LPIPS (0.027), indicating superior reconstruction quality for facial videos. This shows that incorporating both spatial and temporal processing leads to better results compared to 2D-only approaches.

% \noindent\textbf{Only using video dataset}

\noindent\textbf{Face feature Mixing.}
In our experiment, the identity, texture, and attribute weights in the face encoder are set to 1.0, 0.6, and 0.6, respectively. In Figure~\ref{fig:mixing}, we demonstrate the effects of varying texture and attribute weights. As the texture weight increases, we observe an improvement in identity similarity. However, if the texture weight continues to increase, there is a loss in the preservation of the target's attributes (pose and expression). In the extreme case (attribute weight = 1.0, texture weight = 0.2), we find that the model almost simply pastes the source face onto the result.As the attribute weight increases, the target's attributes are better preserved, but identity similarity decreases.

\noindent\textbf{Ablation study.} In the upper part of Figure~\ref{fig:ablation}, we present the results under occlusion conditions. Without the occlusion augmentation, we observe significant distortion of the face, and the occluder (e.g., hand) is either severely deformed or disappears entirely. After applying occlusion data augmentation, the output quality improves dramatically. In the lower part of Figure~\ref{fig:ablation}, we show results for large pose variations. Without the additional guidance from 3D face reconstruction in the denoising network, the generated face becomes highly unstable, with noticeable distortion and deformation.
\section{Conclusion}
\label{sec:conclusion}
\noindent In this paper, we introduced a novel diffusion-based framework for video face swapping, addressing key challenges such as temporal consistency, identity preservation, and handling large pose variations. Our image-video hybrid training strategy leverages both static images and video data, improving model diversity and robustness. The \vaeshortname~, coupled with a custom Attribute-Identity Disentanglement Triplet (AIDT) dataset and 3D Morphable Model integration, enables accurate face swapping while mitigating issues like flickering and occlusions.
Experimental results demonstrate that our framework outperforms existing methods in terms of FVD, temporal consistency, and identity preservation, while requiring fewer inference steps. Overall, our approach provides a more efficient and effective solution for high-quality video face swapping and sets the stage for future advancements in the field.
{
    \small
    \bibliographystyle{ieeenat_fullname}
    \bibliography{main}
}

\end{document}